\documentclass[fleqn,10pt]{SelfArx} 

\usepackage[english]{babel} 

\usepackage{graphicx} 
\usepackage{amsmath,amsfonts,amssymb}
\usepackage{textcomp}
\usepackage{gensymb}
\usepackage{xcolor}
\usepackage{float}
\usepackage{verbatim}
\usepackage[hidelinks]{hyperref}

\newcommand{\norm}[1]{\left\lVert#1\right\rVert}
\graphicspath{{Figures/}}
\usepackage{helvet}

\definecolor{color1}{RGB}{0,0,90} 
\definecolor{color2}{RGB}{0,20,20} 

\usepackage{hyperref} 

\hypersetup{
	hidelinks,
	colorlinks,
	breaklinks=true,
	urlcolor=color2,
	citecolor=color1,
	linkcolor=color1,
	bookmarksopen=false,
	pdftitle={Title},
	pdfauthor={Author},
}

\JournalInfo{Technical Report} 
\Archive{Human-Robot Collaboration and Companionship (HRC$^2$) Lab, Cornell University} 

\PaperTitle{Analytical Inverse Kinematics for a 5-DoF Robotic Arm with a Prismatic Joint} 

\Authors{Vighnesh Vatsal\textsuperscript{1}*, Guy Hoffman\textsuperscript{1}} 
\affiliation{\textsuperscript{1}\textit{Sibley School of Mechanical and Aerospace Engineering, Cornell University, Ithaca, USA}} 
\affiliation{*\textbf{Corresponding author}: vv94@cornell.edu} 

\newcommand{\keywordname}{Keywords} 

\Abstract{We present an analytical solution for the inverse kinematics (IK) of a robotic arm with one prismatic joint and four revolute joints. This 5-DoF design is a result of minimizing weight while preserving functionality of the device in a wearable usage context. Generally, the IK problem for a 5-DoF robot does not guarantee solutions due to the system being over-constrained. We obtain an analytical solution by applying geometric projections and limiting the ranges of motion for each DoF. We validate this solution by reconstructing randomly sampled end-effector poses, and find position errors below 2 cm and orientation errors below 4{\degree}.}

\begin{document}

\maketitle 

\tableofcontents 

\thispagestyle{empty} 


\section{Introduction}
\label{sec:intro}
\begin{figure}[th]
\centering
  \includegraphics[height=0.55\columnwidth]{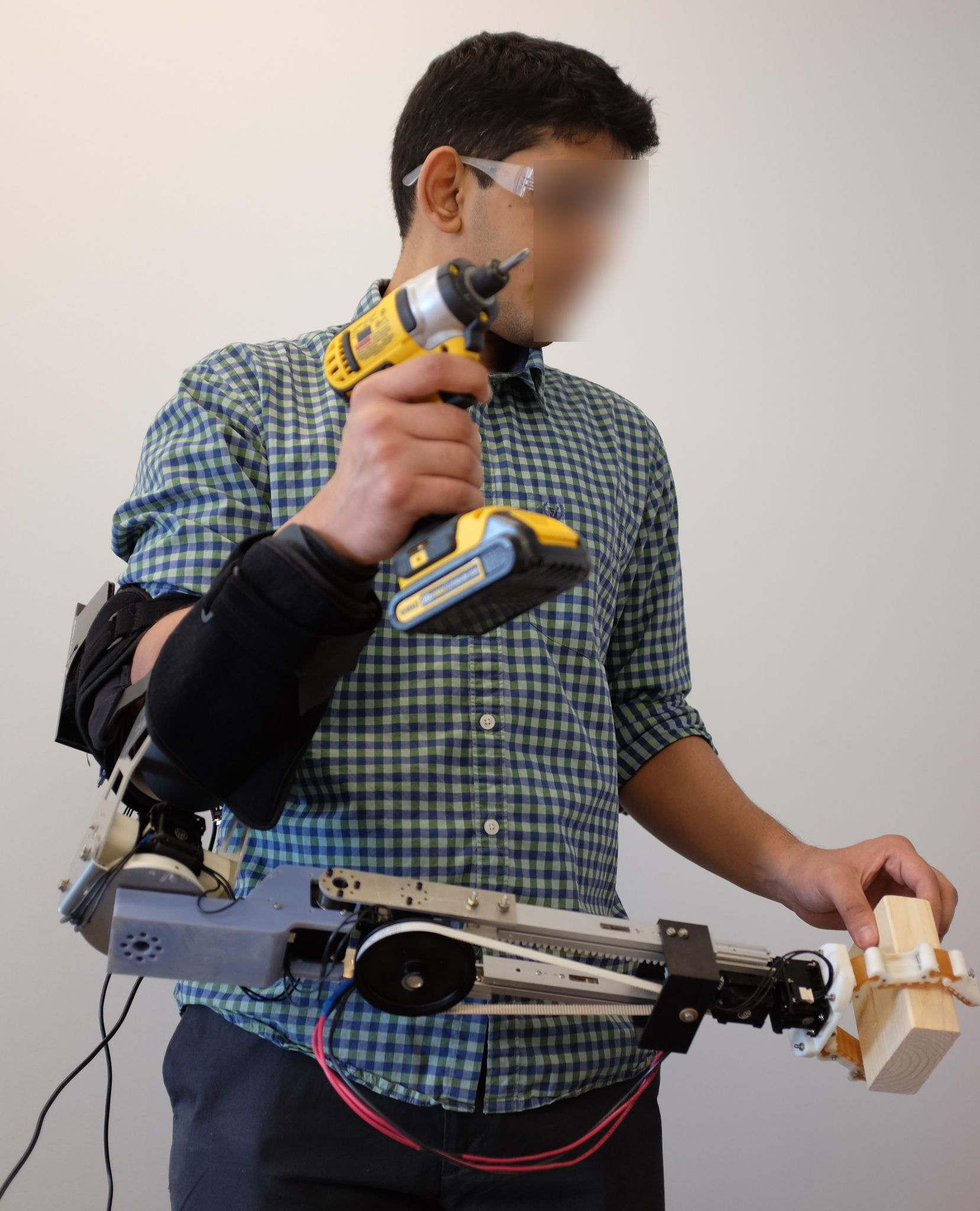}
  \includegraphics[height=0.55\columnwidth]{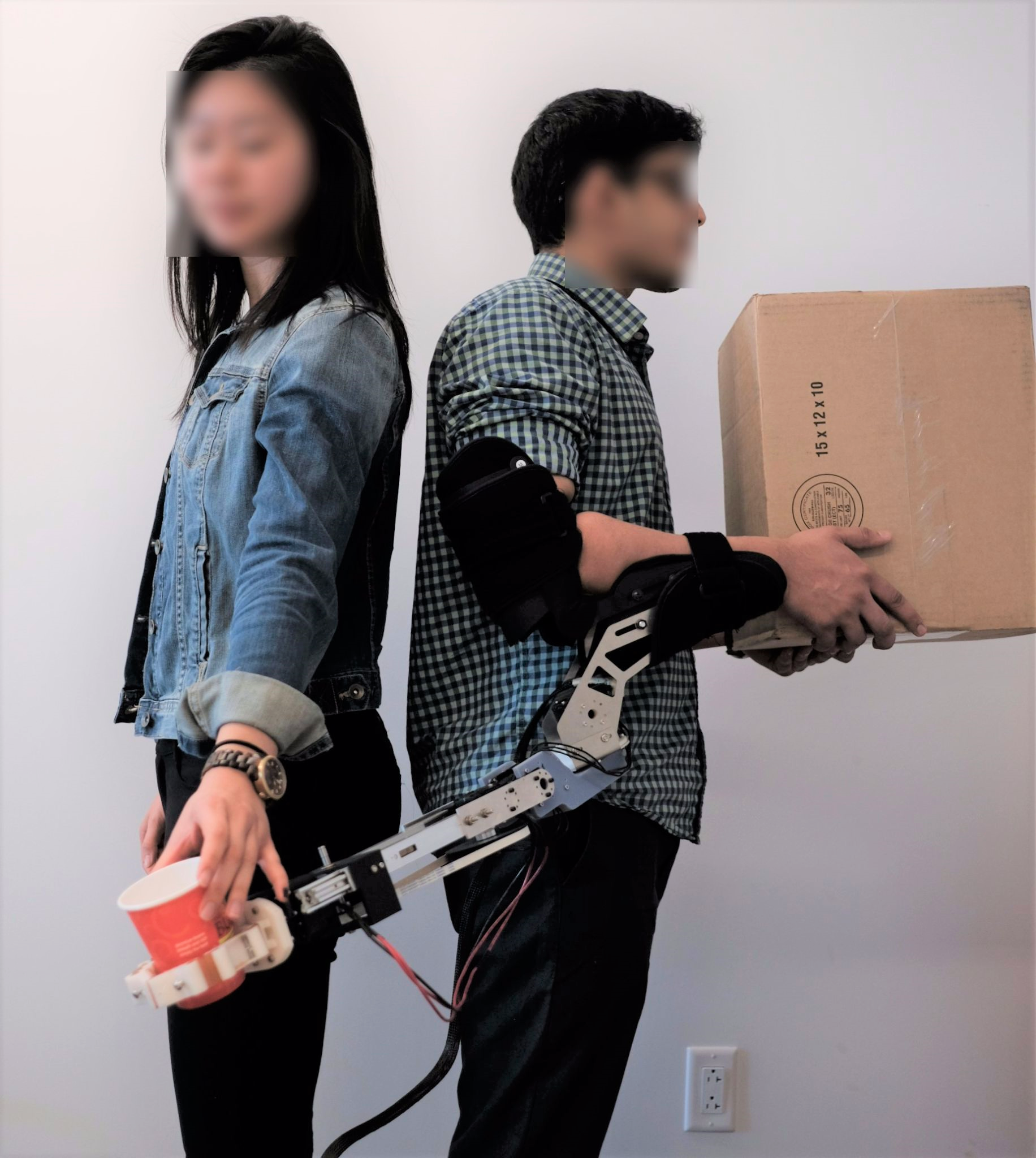}

\caption{The wearable robotic arm in two illustrative usage scenarios: self-handover (left), and handover to another person (right)}
\label{fig:intro}
\end{figure}

\begin{figure}[th]
\centering
  \includegraphics[width=0.93\columnwidth]{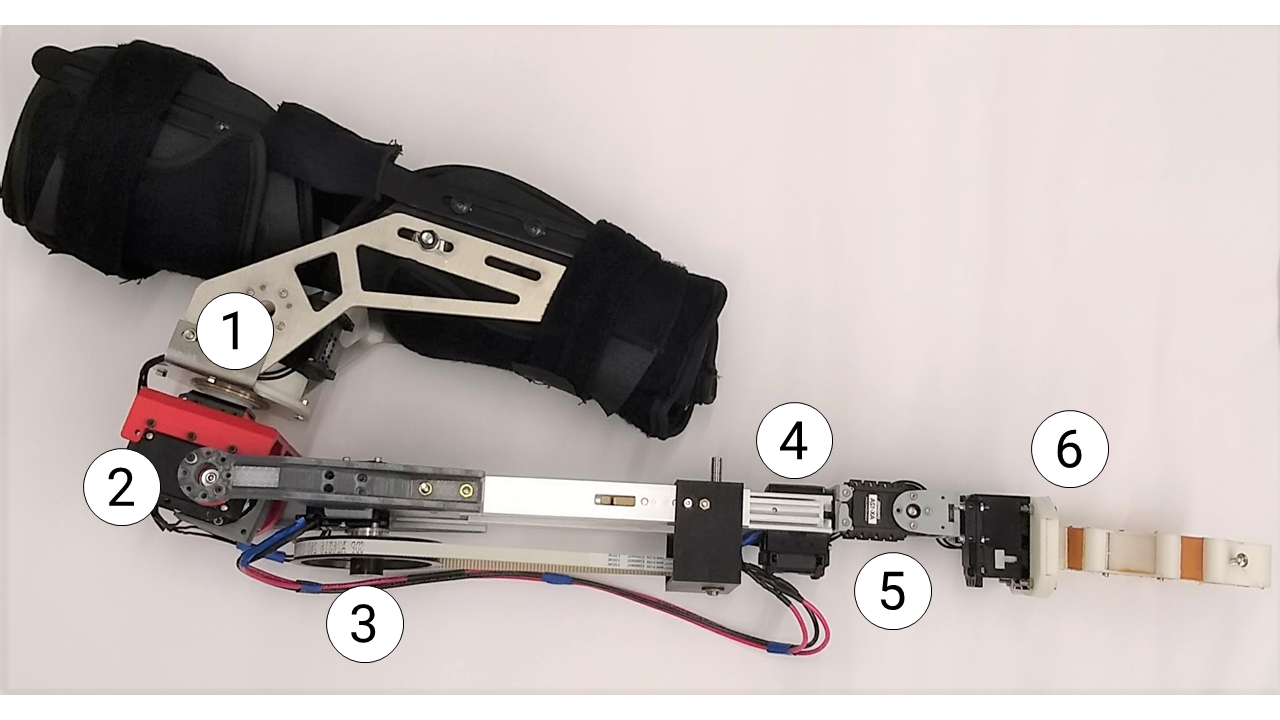}
\caption{Wearable robotic arm with labeled DoFs: 1) Horizontal panning, 2) Vertical pitching, 3) Length extension, 4) Wrist rotation, 5) Wrist pitching, and 6) End-effector}
\label{fig:arm_pic}
\end{figure}

We present an analytical solution for the inverse kinematics (IK) of a robotic arm with five degrees of freedom (DoFs). The kinematic structure of the robot is RRPRR, where ``R" denotes a revolute joint and ``P" denotes a prismatic joint. 
Our use context is the development of a wearable robotic ``third arm" intended for close range human-robot collaboration. It aims to extend a user's reach, and provide the ability for manipulation below and behind the user in tasks similar to those shown in Fig.~\ref{fig:intro}. We find that a 5-DoF structure (excluding gripper) is sufficient for performing these tasks while keeping the weight to a minimum (Fig.~\ref{fig:arm_pic}), as indicated by results from an iterative user-centered design procedure~\cite{vatsal2017wearing}.


Closed-form analytical IK solutions, and numerical solutions exist for general robotic arms with 6-R joints~\cite{lee1982robot, manocha1994efficient, paul1979kinematic}. When possible, analytical methods are preferred over numerical ones due to faster computational speeds and the ability to predict the existence of solutions~\cite{waldron2016kinematics}. 

\begin{figure}[th]
\centering
  \includegraphics[width=0.93\columnwidth]{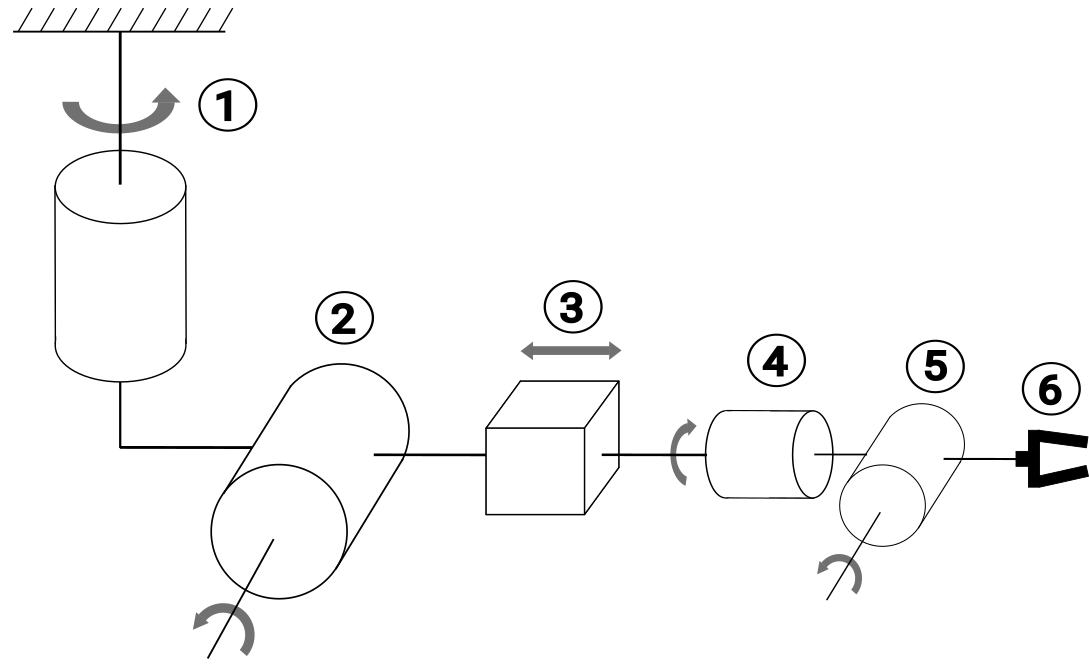}
\caption{Kinematic diagram of the wearable robotic arm.}
\label{fig:kinem_diag}
\end{figure}

However, the IK problem for a 5-DoF robot is over-constrained, and no general solution exists~\cite{tolani2000real}. Gan et al.~\cite{gan2005complete} have developed an analytical IK solution for poses reachable by a 5-R \textit{Pioneer 2} robotic arm. We extend their approach to the case where the robotic arm includes a prismatic joint (Fig.~\ref{fig:kinem_diag}), and present an analytical solution along with the constraints on it.\footnote{The code for this solution can be found here: \url{http://github.com/vighv/5DoF_IK} } We evaluate the numerical quality of this solution by reconstructing a randomly sampled set of end-effector poses within the robot's workspace bounding box. We obtain average position errors below 2 cm, and orientation errors below 4{\degree} in this reconstructed set.  

\begin{table*}[th]

\centering
\caption{D-H parameters for the wearable robotic arm}
\label{tab:dh_v2}
\begin{tabular}{ |l|c|c|c|c| }
  \hline
  \multicolumn{1}{|c|}{Degree of Freedom} & $\alpha_i$ & $a_i$(m) & $d_i$(m) & $\theta_i$\\
  \hline
  1) Horizontal panning & +90\degree & 0 & -0.08 ($l_1$)& (-180{\degree}, 180{\degree})\\
  2) Vertical pitching & +90\degree & 0 & 0 & (0{\degree}, 90{\degree})\\
  3) Length extension & 0\degree & 0 & [0.33, 0.45] & 180{\degree}\\
  4) Wrist rotation & +90\degree & 0 & 0.045 ($l_2$)& (-180{\degree}, 180{\degree})\\
  5) Wrist pitching & +90\degree & 0 & 0 & (0{\degree}, 180{\degree})\\
  6) End-effector & 0\degree & 0.135 ($l_3$)& 0 & 0{\degree}\\
  \hline
 \end{tabular}
\end{table*}

\section{Forward Kinematics Equations}
\label{sec:fk}

Forward Kinematics (FK) provides the mapping from a robot's joint positions to the pose of the end-effector. We assign coordinate frames to each link of the robotic arm, and derive the FK equations using the Denavit-Hartenberg (D-H) convention~\cite{hartenberg1955kinematic}. 
\begin{figure}[th]
\centering
  \includegraphics[width=0.93\columnwidth]{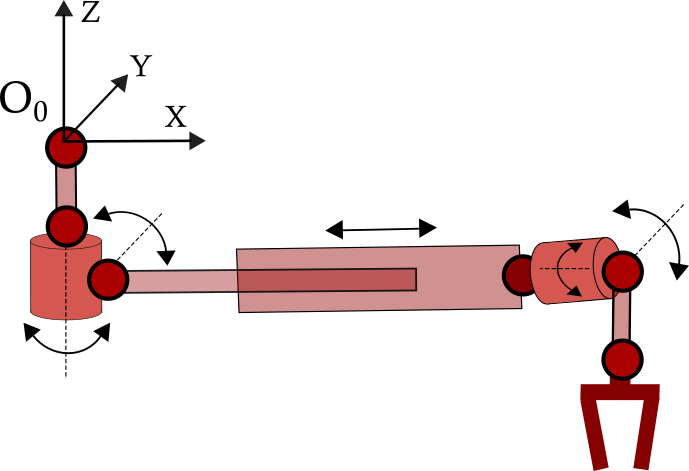}
  \includegraphics[width=0.93\columnwidth]{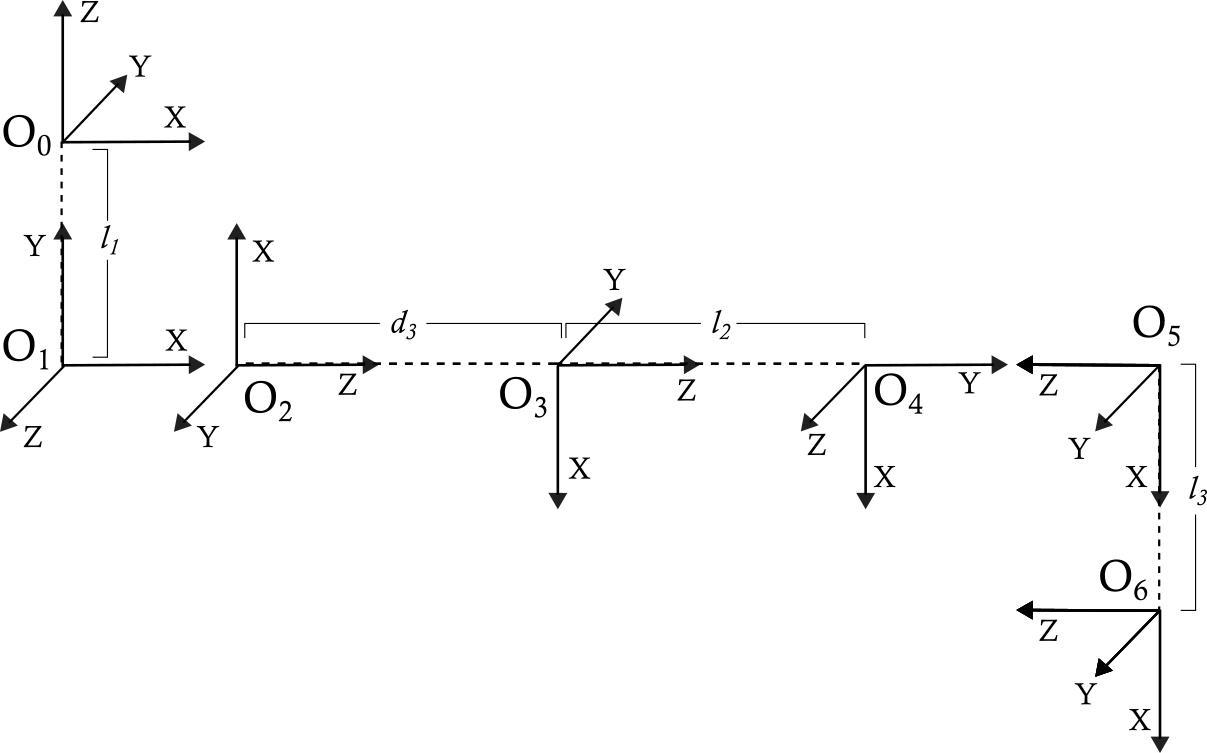}
\caption{Schematic of the robotic arm and associated coordinate frames: $O_0$ is the base frame, $O_6$ is the end-effector. Frame $O_1$ is coincident with $O_2$, and $O_4$ is coincident with $O_5$.}
\label{fig:coord}
\end{figure}
 
The base frame $O_0$ in our case lies at the top of horizontal panning DoF (Fig.~\ref{fig:coord}), and the final frame $O_6$ is attached to the end-effector, lying at the mid point of the gripper's fingers. The D-H method uses four parameters: $\alpha, a, d $ and $\theta$ to specify the relationship between frames (Table~\ref{tab:dh_v2}). Using these parameters, a homogeneous transformation matrix $T^{i+1}_i$ is defined for the pose of frame $i+1$ as seen in frame $i$:
\begin{equation}
T^{i+1}_i = \begin{bmatrix}
cos \theta_i & -sin \theta_i cos \alpha_i & sin \theta_i sin \alpha_i & a_i cos \theta_i \\
sin \theta_i & cos \theta_i cos \alpha_i & -cos \theta_i sin \alpha_i & a_i sin \theta_i \\
0 & sin \alpha_i & cos \alpha_i & d_i \\
0 & 0 & 0 & 1
\end{bmatrix}
\end{equation}

The joint variables are $\theta_i$ for the four revolute joints, and $d_3$ for the prismatic joint. The end-effector is fixed with respect the wrist pitching DoF. However, it is listed separately in order to decouple it from wrist pitching for ease of solution of the inverse kinematics, as in~\cite{xu2005analysis}. 

The transformation $T^6_0$, between the base frame $O_0$ and the end-effector frame $O_6$ can be found by successive multiplication:

\begin{equation}
\label{eqtran}
T^6_0 = T^1_0 T^2_1 T^3_2 T^4_3 T^5_4 T^6_5
\end{equation}


Expanded out in terms of individual elements:
\begin{equation}
T^6_0 = \begin{bmatrix}
R_{1x} & R_{2x} & R_{3x} & P_x \\
R_{1y} & R_{2y} & R_{3y} & P_y \\
R_{1z} & R_{2z} & R_{3z} & P_z \\
0 & 0 & 0 & 1
\end{bmatrix}
\end{equation}

The rotation between these frames is described by $R = [\vec{R}_1, \vec{R}_2, \vec{R}_3] \in SO(3)$, where $SO(3)$ is the 3D rotation group consisting of 3$\times$3 orthogonal matrices with unit determinant. $\vec{P} = [P_x, P_y, P_z]^T \in \mathbb{R}^3$  describes the translation between these frames. We will denote $cos \theta_i$ as $c_i$ and $sin \theta_i$ as $s_i$ in the rest of this paper. The length parameters for the robot are $l_1$, $l_2$ and $l_3$, which correspond to $d_1$, $d_4$ and $a_6$ respectively from Table~\ref{tab:dh_v2}. There are twelve equations that constitute the forward kinematics:

\begin{equation}
\label{eq1}
R_{1x} = c_1 s_2 s_5 - c_5 (s_1 s_4 \ + \ c_1 c_2 c_4)
\end{equation}
\begin{equation}
\label{eq2}
R_{1y} = c_5 (c_1 s_4 \ - \ c_2 c_4 s_1) + s_1 s_2 s_5
\end{equation}
\begin{equation}
\label{eq3}
R_{1z} = -c_2 s_5 - c_4 c_5 s_2 
\end{equation}
\begin{equation}
\label{eq4}
R_{2x} = -c_4 s_1 - c_1 c_2 s_4
\end{equation}
\begin{equation}
\label{eq5}
R_{2y} = -c_1 c_4 - c_2 s_1 s_4
\end{equation}
\begin{equation}
\label{eq6}
R_{2z} = -s_2 s_4
\end{equation}
\begin{equation}
\label{eq7}
R_{3x} = -c_1 s_2 c_5 - s_5 (s_1 s_4 \ + \ c_1 c_2 c_4)
\end{equation}
\begin{equation}
\label{eq8}
R_{3y} = s_5 (c_1 s_4 \ - \ c_2 c_4 s_1) - s_1 s_2 c_5
\end{equation}
\begin{equation}
\label{eq9}
R_{3z} = c_2 c_5  - s_2 c_4 s_5
\end{equation}
\begin{equation}
\label{eq10}
P_{x} = l_2 c_1 s_2 - l_3(c_5 s_1 s_4 \ + \ c_5 c_1 c_2 c_4) + l_3 c_1 s_2 s_5 + d_3 c_1 s_2
\end{equation}
\begin{equation}
\label{eq11}
P_{y} = l_2 s_1 s_2 + l_3(c_5 c_1 s_4 \ + \ c_5 s_1 c_2 c_4) + l_3 s_1 s_2 s_5 + d_3 s_1 s_2
\end{equation}
\begin{equation}
\label{eq12}
P_{z} = l_1 - l_2 c_2 - l_3 c_2 s_5 - l_3 c_4 c_5 s_2 - d_3 c_2
\end{equation}

\section{Inverse Kinematics}
\label{sec:ik}

The inverse kinematics problem involves finding the values of the joint variables for a desired position and orientation (pose) of the end-effector. 

\begin{figure}[th]
  \includegraphics[width=0.93\columnwidth]{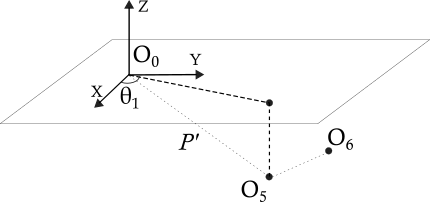}
\caption{Projecting the position of $O_5$ onto the $XY$-plane of the base frame $O_0$ to solve for $\theta_1$.}
\label{fig:theta1}
\end{figure}

We start with the solution to the horizontal panning angle $\theta_1$ using the geometric projection method described in~\cite{xu2005analysis} and~\cite{gan2005complete}. As shown in Fig.~\ref{fig:theta1}, $\theta_1$ is the angle between the projection of $\vec{P'} = \vec{O_0 O_5}$ onto the $XY$-plane, and the $X$-axis of the base frame $O_0$. $\vec{P'}$ can be obtained by using the transformation between $O_5$ and $O_0$:

\begin{equation}
T^5_0 = T^6_0  \ [ T^6_5 ]^{-1} = \begin{bmatrix}
\vec{R'}_1 & \vec{R'}_2 & \vec{R'}_3 & \vec{P'} \\
0 & 0 & 0 & 1
\end{bmatrix}
\label{eq_geom}
\end{equation}

The components $P'_y$ and $P'_x$ of $\vec{P'}$ can be found from the corresponding components in the last column of $T^5_0$. The relation between $\theta_1$ and the projections of $\vec{P'}$ is as follows:

\begin{equation}
tan \theta_1 = \frac{P'_y}{P'_x} = \frac{-l_3 R_{1y} + P_y}{-l_3 R_{1x} + P_x}
\end{equation} 

$\theta_1$ can be found using the four-quadrant inverse tangent function $atan2(x,y)$, as it lies in the desired range of (-180{\degree}, 180{\degree}). Also note that $\theta_2$ is never fully vertical, since we have restricted the range of DoF-2 in the open interval (0{\degree}, 90{\degree}). This ensures that a solution for $\theta_1$ always exists, since $O_0$, $O_1$, $O_2$ and $O_5$ are never collinear.

The solution strategies for other joint variables can be grouped into two cases, depending on whether $R_{2z} \neq 0$ or $R_{2z} = 0$, which correspond to whether the links of the robot become coplanar.

\subsection{Case 1: $R_{2z} \neq 0$}
\label{subsec:case1}

From the forward kinematics, we know that:
\begin{equation}
R_{2z} = - s_2 s_4
\end{equation}

$R_{2z} \neq 0$ implies that both $s_2 \neq 0$ and $s_4 \neq 0$. Then we can perform the following substitutions: 

\begin{equation}
s_4 = \frac{-R_{2z}}{s_2}
\end{equation} 

This results in a pair of linear equations in $cos \theta_4$ and $cot \theta_2$:

\begin{equation}
\begin{bmatrix}
s_1 & c_1 R_{2z} \\
-c_1 & s_1 R_{2z}
\end{bmatrix}
\begin{bmatrix}
cos \theta_4 \\
cot \theta_2
\end{bmatrix} = 
\begin{bmatrix}
R_{2x} \\
R_{2y}
\end{bmatrix}
\end{equation}

We can find $tan \theta_2$ by solving this pair of linear equations, and constrain the solution to lie in the range (0{\degree}, 90{\degree}). These equations also give us $cos \theta_4$, with which we can find $tan \theta_4$:

\begin{equation}
tan \theta_4 = \frac{-R_{2z}}{s_2 \ cos \theta_4}
\end{equation}

$\theta_4$ is computed using the four-quadrant inverse tangent function $atan2(x,y)$ as it lies in the range (-180{\degree}, 180{\degree}). An inverse tangent function is preferred over an inverse cosine function due to greater numerical stability. Once $\theta_1$, $\theta_2$ and $\theta_4$ are known, equations (\ref{eq3}) and (\ref{eq9}) reduce to linear expressions in terms of $sin \theta_5$ and $cos \theta_5$:

\begin{equation}
\begin{bmatrix}
-c_2 & -c_4 s_2 \\
-c_4 s_2 & c_2
\end{bmatrix}
\begin{bmatrix}
sin \theta_5 \\
cos \theta_5
\end{bmatrix} = 
\begin{bmatrix}
R_{1z} \\
R_{3z}
\end{bmatrix}
\end{equation}

$\theta_5$ can be obtained using the inverse tangent function $atan2(x,y)$, and constrained to lie in the range (0{\degree}, 180{\degree}). Once all joint angles are known, equations (\ref{eq10}-\ref{eq12}) yield three candidate solutions for the length of the prismatic joint $d_3$:

\begin{equation}
d_3^1 = (P_x - l_2 c_1 s_2 + l_3(c_5 s_1 s_4 \ + \ c_5 c_1 c_2 c_4) - l_3 c_1 s_2 s_5) / c_1 s_2
\label{eqd31}
\end{equation}
\begin{equation}
d_3^2 = (P_y - l_2 s_1 s_2 - l_3(c_5 c_1 s_4 \ + \ c_5 s_1 c_2 c_4) - l_3 s_1 s_2 s_5) / s_1 s_2
\label{eqd32}
\end{equation} 
\begin{equation}
d_3^3 = (P_z - l_1 + l_2 c_2 + l_3 c_2 s_5 + l_3 c_4 c_5 s_2) / c_2
\label{eqd33}
\end{equation}    

The candidate for $d_3$ that lies within the range [0.33 m, 0.45 m] is chosen as the solution.

\subsection{Case 2: $R_{2z} = 0$}
\label{subsec:case2}

A different strategy must be employed in situations when $R_{2z}$ is zero. Since $R_{2z} = s_2 s_4$, and we have restricted $\theta_2$ to lie in the open interval (0{\degree}, 90{\degree}), $R_{2z} = 0 \Rightarrow s_4 = 0$. Additionally, since $\theta_4$ is also restricted to be in the open interval (-180{\degree}, 180{\degree}), we obtain $\theta_4 = 0$.

\begin{figure}[th]
\centering
  \includegraphics[width=0.93\columnwidth]{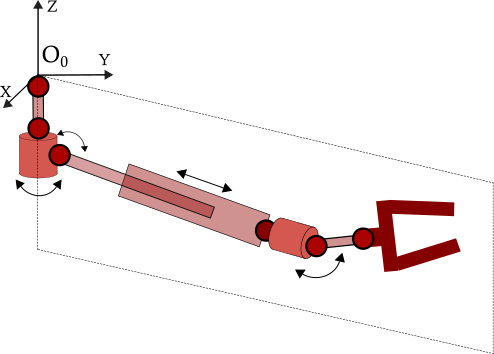}
\caption{All links of the robotic arm become co-planar when $\theta_4 = 0$.}
\label{fig:case2}
\end{figure}

This places the robot in a configuration similar to the one shown in Fig.~\ref{fig:case2}, where all the links of the arm become co-planar. With $\theta_4 = 0$, equations (\ref{eq1}, \ref{eq2}) and (\ref{eq7} - \ref{eq9}) simplify to:

\begin{equation}
R_{1x} = -c_1 cos(\theta_2 + \theta_5)
\end{equation}
\begin{equation}
R_{1y} = -s_1 cos(\theta_2 + \theta_5)
\end{equation}
\begin{equation}
R_{3x} = -c_1 sin(\theta_2 + \theta_5)
\end{equation}
\begin{equation}
R_{3y} = -s_1 sin(\theta_2 + \theta_5)
\end{equation}
\begin{equation}
R_{3z} = cos(\theta_2 + \theta_5)
\end{equation}

This gives us the sum $(\theta_2 + \theta_5)$:

\begin{equation}
tan(\theta_2 + \theta_5) = \frac{R_{3x}}{R_{1x}} = \frac{R_{3y}}{R_{1y}}
\end{equation}

\begin{figure}[th]
\centering
  \includegraphics[width=0.8\columnwidth]{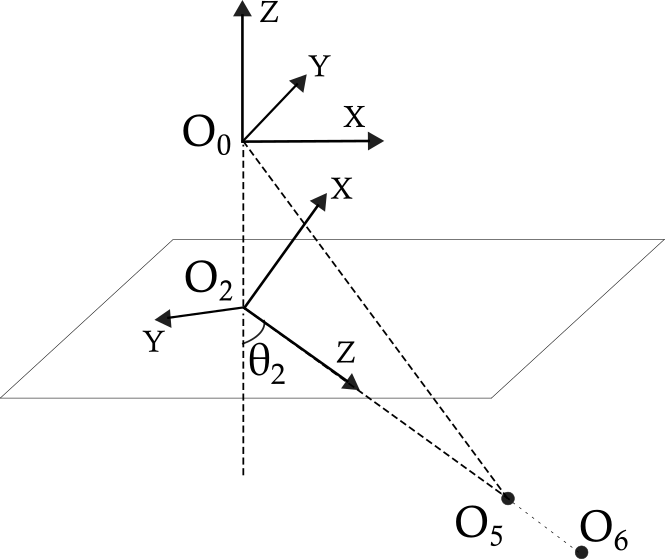}
  \includegraphics[width=0.85\columnwidth]{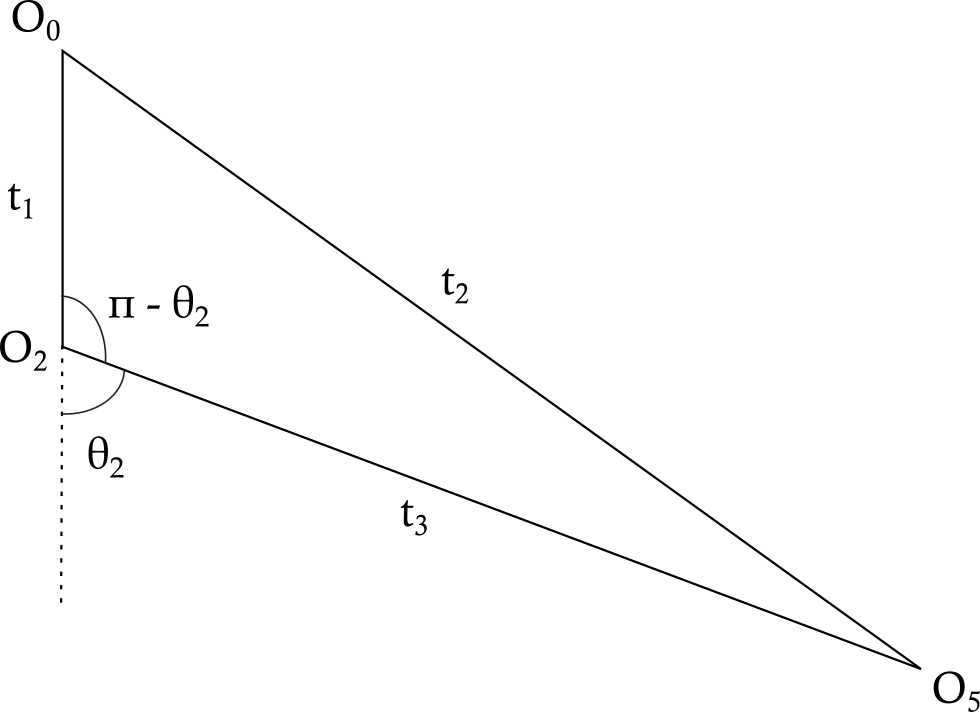}
\caption{$O_0$, $O_2$ and $O_5$ are coplanar. The triangle formed by these points is used to find $\theta_2$.}
\label{fig:c2thet2}
\end{figure}

To find the the joint angle $\theta_2$ separately, we look at the triangle formed by the points $O_0$, $O_2$ and $O_5$, as shown in Fig.~\ref{fig:c2thet2}. 

The edge lengths $t_1$, $t_2$ and $t_3$ of this triangle are known quantities, since the vector $\vec{O_0 O_2}$ is fixed, and the vector $\vec{O_0 O_5}$ can be found using the last column of the matrix $T^5_0$ in equation (\ref{eq_geom}). We can apply the cosine formula to this triangle:

\begin{equation}
t_1 = \norm{\vec{O_0 O_2}}
\end{equation}

\begin{equation}
t_2 = \norm{\vec{O_0 O_5}}
\end{equation}

\begin{equation}
t_3 = \norm{\vec{O_0 O_2} - \vec{O_0 O_5}}
\end{equation}

\begin{equation}
cos(\pi - \theta_2) = \frac{t_1^2 + t_3^2 - t_2^2}{2 t_1 t_3}
\end{equation}

We can now solve for $\theta_2$:

\begin{equation}
\theta_2 = \pi - acos([\frac{t_1^2 + t_3^2 - t_2^2}{2 t_1 t_3}])
\end{equation}

Once $\theta_2$ is known, we get two possible candidates for $\theta_5$:

\begin{equation}
\theta_5 = tan^{-1} \frac{R_{3x}}{R_{1x}} - \theta_2 = tan^{-1} \frac{R_{3y}}{R_{1y}} - \theta_2
\end{equation}

Having found all the joint angles, we can proceed as in Section~\ref{subsec:case1}, using equations (\ref{eqd31} - \ref{eqd33}) to find $d_3$, the length of the prismatic joint.

\section{Constraints on Solutions}
\label{sec:constraints}
In Sections~\ref{subsec:case1} and~\ref{subsec:case2}, we chose between multiple candidate solutions by applying the condition that the variables must lie within the intervals specified in Table~\ref{tab:dh_v2}. Apart from these, there are also conditions on the revolute joint angles $\theta_1$ - $\theta_5$ that arise from the forward kinematic equations. These conditions make use of the individual transformation matrices $T^{i+1}_i$ for each row of Table~\ref{tab:dh_v2}.

%
%
%
%
%
%
Let us consider the product $T^1_0 T^2_1 T^3_2$:
\begin{equation}
T^1_0 T^2_1 T^3_2 = \begin{bmatrix}
-c_1 c_2 & -s_1 & c_1 s_2 & d_3 c_1 s_2 \\
-c_2 s_1 & c_1 & s_1 s_2 & d_3 s_1 s_2 \\
-s_2 & 0 & -c_2 & l_1 - d_3 c_2 \\
0 & 0 & 0 & 1
\end{bmatrix}
\end{equation}

The entry in row 3, column 2 of this matrix is zero. Using equation (\ref{eqtran}), we can compute this matrix in terms of the final transformation $T^6_0$ :

\begin{equation}
T^1_0 T^2_1 T^3_2 = T^6_0 [T^4_3 T^5_4 T^6_5]^{-1}
\end{equation} 

Equating corresponding matrix entries at row 3, column 2 on both sides, we obtain the following condition on $\theta_4$ and $\theta_5$:
\begin{equation}
\label{invar1}
R_{1z} c_5 s_4 - R_{2z} c_4 + R_{3z} s_4 s_5 = 0
\end{equation}

Similarly, in the product $T^4_3 T^5_4 T^6_5$, the element at row 3, column 2 is also zero:

\begin{equation}
T^4_3 T^5_4 T^6_5 = \begin{bmatrix}
c_4 c_5 & s_4 & c_4 s_5 & l_3 c_4 c_5 \\
s_4 c_5 & -c_4 & s_4 s_5 & l_3 s_4 c_5 \\
s_5 & 0 & -c_5 & l_2 + l_3 s_5 \\
0 & 0 & 0 & 1
\end{bmatrix}
\end{equation}

This can also be written in terms of the final transformation matrix:
\begin{equation}
T^4_3 T^5_4 T^6_5 = [T^1_0 T^2_1 T^3_2]^{-1} T^6_0 
\end{equation} 

Similar to equation (\ref{invar1}), we obtain the following condition on $\theta_1$ and $\theta_2$:
\begin{equation}
\label{invar2}
R_{2x} c_1 s_2 - R_{2z} c_2 + R_{2y} s_1 s_2 = 0
\end{equation}

Candidate IK solutions that do not satisfy conditions (\ref{invar1}) and (\ref{invar2}) are rejected. 


\section{Evaluation}
\label{sec:sim}

The analytical IK solution presented here is exact. It should yield no errors if the computations are performed with infinite numerical precision. However, the inverse tangent and inverse cosine functions used in our formulation are numerically sensitive, and are susceptible to propagation of errors~\cite{gan2005complete}.

To evaluate the quality of our solution, 
we follow the procedure by which a robotic arm is generally actuated, where the joint angles for a desired pose are computed using IK and sent to the motors, resulting in an end-effector pose through FK. Our evaluation measures the numerical errors that will manifest in a physical implementation of this pose reconstruction, even with the assumption of arbitrarily precise actuators.

\begin{figure}[th]
\centering
  \includegraphics[width=0.75\columnwidth]{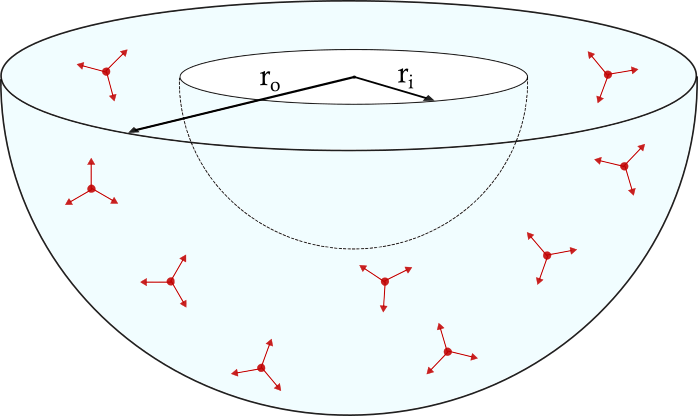}  
\caption{Schematic of the sampling process for IK evaluation: 3-D poses are sampled from within a hemispherical volume, lying between radii $r_i$ and $r_o$, that bounds the robot's workspace.}
\label{fig:sample_schematic}
\end{figure}

The bounding volume of the robot's workspace lies in a hemispherical shell of inner radius $r_i = d_3^{min} + l_2$ and outer radius $r_o = d_3^{max} + l_2 + l_3$. We sample end-effector positions from uniform distributions in $X$, $Y$ and $Z$ coordinate ranges within this volume. Orientations are sampled as Yaw-Pitch-Roll Euler angles from uniform distributions in the range [0, 2$\pi$] for yaw and roll, and [0, $\pi$] for pitch (Fig.~\ref{fig:sample_schematic}).

Each sampled pose is represented by a position vector $\vec{P}$, and a rotation matrix $R$ constructed from these Euler angles. From the IK computation on this pose, we obtain a vector $\vec{\Theta}$ of joint variables. This pose is reconstructed into a position $\vec{P}'$ and rotation matrix $R'$ on applying FK equations to $\vec{\Theta}$. We consider the absolute error as the accuracy metric for position:

\begin{equation}
\begin{bmatrix}
\Delta X_{abs} \\
\Delta Y_{abs} \\
\Delta Z_{abs} 
\end{bmatrix} =  \begin{bmatrix}
| P_x - P'_x | \\
| P_y - P'_y | \\
| P_z - P'_z |
\end{bmatrix}
\end{equation}

For the accuracy of rotation reconstruction, we consider the geodesic metric on the rotation group $SO(3)$, as described in~\cite{huynh2009metrics}:

\begin{equation}
\Delta R_{abs} = || log(R' R^{T}) ||
\end{equation}

Where the norm of a matrix $A$ is defined as:
\begin{equation}
|| A || = \sqrt{\frac{1}{2} \ trace (A^T A)}
\end{equation}

This metric returns the magnitude of rotation angle in the range [0,$\pi$) measured along the geodesic between $R$ and $R'$ in $SO(3)$. It captures rotational deviation more accurately compared to the Euclidean distance between Euler angles, since a smaller Euclidean distance can map to a larger distance in $SO(3)$ and vice versa~\cite{huynh2009metrics}. 

We sampled $10^7$ poses in the robot's bounding workspace volume, out of which $N = 7.4 \times 10^6$ poses satisfied the constraint equations (\ref{invar1}) and (\ref{invar2}), and were within the range of prismatic joint lengths.  Table~\ref{tab:bb_sim_result} lists the means and standard deviations of the absolute position and orientation errors for the reconstruction of these poses. The mean position errors are in the range of $\sim$ 6 mm in the $X$ and $Y$ coordinates, and $\sim$ 2 cm in the $Z$ coordinate. These errors respectively correspond to 1.18\% and 3.92\% of the arm's total maximum length (0.51 m). The mean error in the $Z$ coordinate is higher than in $X$ and $Y$, possibly due to the greater sensitivity of $P_z$ to errors in the estimate of the prismatic joint length $d_3$ (equation~\ref{eq12}). The mean orientation error is $\sim$ 0.054 radians, or just above 3{\degree}. 

\begin{table}

\centering
\caption{Absolute errors in pose reconstruction}
\label{tab:bb_sim_result}
\begin{tabular}{ |c|c|c| }
  \hline
  \multicolumn{1}{|c|}{Quantity} & Mean & Std. Dev. \\
  \hline
  $\Delta X_{abs}$ (m) & 6.099 $\times 10^{-3}$ & 6.543 $\times 10^{-3}$ \\
  $\Delta Y_{abs}$ (m) & 6.092 $\times 10^{-3}$ & 6.554 $\times 10^{-3}$ \\
  $\Delta Z_{abs}$ (m) & 1.681 $\times 10^{-2}$ & 1.448 $\times 10^{-2}$ \\
  $\Delta R_{abs}$ (rad) & 5.343 $\times 10^{-2}$ & 1.972 $\times 10^{-1}$\\
  \hline
 \end{tabular}
\end{table}

\section{Conclusion}
\label{sec:conclusion}
We presented an analytical solution for the inverse kinematics of a 5-DoF wearable robotic arm with a prismatic actuator present in the serial chain. We validated our solution by testing it on $10^7$ randomly generated poses, and found that we were able to achieve orientation targets to within 4{\degree}, and position targets to within 2 cm in terms of numerical errors. We also derived constraints on the joint variables to verify the correctness of solutions for a particular end-effector pose.

Although this method returns solutions for poses that lie within the subspace reachable by our 5-DoF arm, it is numerically unstable for poses that lie near the ends of the joint variable ranges. On the physical robot, we would require approximate solutions for planned end-effector trajectories that may contain poses near these singular points. In such cases, we can augment our analytical solution with a numerical approach, such as gradient descent-based search~\cite{xu2005analysis} to obtain a more stable, hybrid solution. 

In the context of developing control methods for the wearable robotic arm, the analytical inverse kinematics solution presented here will reduce the computational costs compared to a numerical strategy, leading to better responsiveness of the robot to the user.


\phantomsection
\section*{Acknowledgments} 

\addcontentsline{toc}{section}{Acknowledgments} 

This work was supported by the National Science Foundation under NRI Award no. 1734399.


\phantomsection
\bibliographystyle{unsrt}
\bibliography{refs}


\end{document}